\documentclass{llncs}
\usepackage{graphicx}
\usepackage{url}
\usepackage{paralist}
\usepackage{soul, color}
\sethlcolor{green}
\usepackage{amssymb}

 \usepackage{algorithmicx}
 \usepackage{algpseudocode}

\begin{document}

\title{Test-Driven Development of ontologies (extended version)}
\author{C. Maria Keet\inst{1} and Agnieszka \L{}awrynowicz\inst{2}}
\institute{Department of Computer Science, University of Cape Town, South Africa
            \email{mkeet@cs.uct.ac.za}
            \and
            Institute of Computing Science, Poznan University of Technology, Poland \email{agnieszka.lawrynowicz@cs.put.poznan.pl}}
\maketitle              


\begin{abstract}
Emerging ontology authoring methods to add knowledge to an ontology focus on ameliorating the validation bottleneck. The verification of the newly added axiom is still one of trying and seeing what the reasoner says, because a systematic testbed for ontology authoring is missing. We sought to address this by introducing the approach of test-driven development for ontology authoring. We specify 36 generic tests, as TBox queries and TBox axioms tested through individuals, and structure their inner workings in an `open box'-way, which cover the OWL 2 DL language features. This is implemented as a Protege plugin so that one can perform a TDD test as a black box test. We evaluated the two test approaches on their performance. The TBox queries were faster, and that effect is more pronounced the larger the ontology is. We provide a general sequence of a TDD process for ontology engineering as a foundation for a TDD methodology.
\end{abstract}

\section{Introduction}
\label{sec:intro}

The process of ontology development has progressed much over the past 20 years, aided by development information systems-oriented methodologies, such as {\sc methontology} for a single-person/group monolithic ontology to ontology networks with NeOn \cite{Suarez08}, and both stand-alone and collaborative tools, such as Prot\'eg\'e \cite{Gennari03} and MoKI \cite{Ghidini09}. They are examples of generic, high-level information systems-oriented methodologies, but support for effective {\em ontology authoring}---adding the right axioms and adding the axioms right---has received some attention only more recently. Processes at this `micro' level of the development process, rather than the `macro' level, may use the reasoner to propose axioms with FORZA \cite{KKG13forza}, use Ontology Design Patterns (ODPs) \cite{Gangemi09}, and start gleaning ideas from software engineering practices, notably exploring the notion of unit tests \cite{Vrandecic06}, eXtreme Design with ODPs \cite{Blomqvist12}, and Competency Question (CQ)-based authoring using SPARQL \cite{Ren14}. However, testing whether a CQ can be answered does not say how to add/change the knowledge represented in the ontology, FORZA considers simple object properties only, and eXtreme Design limits one to ODPs that have to come from some place. Put differently, there is no systematic testbed for ontology engineering, to implement the CQ in the authoring process in a piecemeal fashion, other than manual efforts by a knowledge engineer to add or change something and running the reasoner to check its effects. This still puts a high dependency on expert knowledge engineering, which ideally should not be in the realm of an art, but rather at least a systematic process for good practices. 

We aim to address this problem by borrowing another idea from software engineering: {\em test-driven development} (TDD). TDD ensures that what is added to the program core (here: ontology) does indeed have the intended effect specified upfront. Moreover, TDD in principle is cognitively a step up from the `just add stuff and lets see what happens'-attitude, therewith deepening the understanding of the ontology authoring process and the logical consequences of an axiom. In addition, it would make roll-backs and conflicting CQs easier to manage. At an informal, high level, one can specify the following three scenarios of usage.
\begin{compactenum}
\item[{\em I. CQ-driven TDD}]
Developers (domain experts, knowledge engineers etc) specify CQs. A CQ is translated automatically into one or more axioms. This (these) axiom(s) are the input of the relevant TDD test(s) to be carried out. The developers who specify the CQs could be oblivious to the inner workings of the two-step process of translating the CQ and testing the axiom(s). 

\item[{\em II-a. Ontology authoring-driven TDD - the knowledge engineer}] The knowledge \linebreak engineer knows which axiom s/he wants to add, types it, which is then fed directly into the TDD system.		

\item[{\em II-b. Ontology authoring-driven TDD - the domain expert}] As there is practically only a limited amount of `types' of axioms to add, one could create templates, alike the notion of the ``logical macro'' ODP \cite{Presutti08}. For instance, a domain expert could choose the all-some template from a list, which then in the TDD system amounts to an axiom of the form $C \sqsubseteq \exists R.D$. The domain expert instantiates it with  relevant domain entities (e.g., ${\sf Professor \sqsubseteq \exists teaches.Course}$), and the related TDD test is then run automatically. The domain expert need not necessarily know the logic, but behind the usability interface, what gets sent to the TDD system is that axiom.									
 \end{compactenum}
While in each case the actual testing can be hidden from the user's view, it is necessary to specify what actually happens during such testing and how it is tested (in a similar way that it needed to be clear how the OWL reasoner works). Here, we assume that either the first step of the CQ process is completed, or the knowledge engineer adds the axiom, or that the template is populated, respectively; i.e., that we are at the stage where the axioms are fed into the TDD test system. To realise the testing, a number of questions have to be answered:
\begin{compactenum}
\item Given the TDD procedure in software engineering---check that desired feature is absent, code it, test again---then what does that mean for ontology testing when transferred to ontology development? 
\item TDD requires so-called {\em mock objects} for `incomplete' parts of the code, and mainly for methods; is there a parallel to it in ontology development, or can that aspect of TDD be ignored?
\item In what way, and where, (if at all) can this be integrated as a methodological step in existing ontology engineering methodologies that are typically based on waterfall, iterative, or lifecycle principles rather than agile methodologies?
\end{compactenum}
%

To work this out for ontologies, we take some inspiration from TDD for conceptual modelling. Tort et al. \cite{Tort11} essentially specify `unit tests' for each feature/possible addition to a conceptual model, and test such an addition against sample individuals. Translating this to OWL ontologies, such testing is possible by means of ABox individuals, and then instead if using an ad hoc algorithm, one can avail of the automated reasoner. In addition, for ontologies, one can avail of a query language for the TBox, namely, SPARQL-OWL \cite{Kollia11}, and most of the test can be specified in that language as well. We define TBox and ABox-driven TDD tests for the basic axioms one can add to an OWL 2 DL ontology. Mock objects are required especially for TDD tests for the `RBox', i.e., for the object property-specific axioms. To examine practical feasibility for the ontology engineer and determine which TDD strategy is the best option for ontology engineering, we first implemented this by means of a Prot\'eg\'e plugin as basic interface, and, second, evaluated the plugin on performance by comparing TBox and ABox TDD tests. TBox queries generally outperform the ABox ones, and this difference is more pronounced with larger ontologies. Finally, we outline a TDD ontology development methodology, which, while having overlap with some extant methodologies, cannot not neatly be squeezed into them. Overall, we thus add a new mechanism and tool to the ontology engineer's `toolbox' to enable systematic development of ontologies in an agile way.

The remainder of the paper is structured as follows. Section~\ref{sec:relwork} describes related works on TDD in software and ontology development. Section~\ref{sec:core} summarises the TDD tests and Section~\ref{sec:eval} evaluates them on performance with the Prot\'eg\'e plugin. 
We discuss in Section~\ref{sec:disc}, where we also answer the above-mentioned research questions, and conclude in Section~\ref{sec:concl}.

%

\section{Related work}
\label{sec:relwork}

Before assessing related works in ontology engineering, we first describe some pertinent aspects of TDD from software engineering.

\subsubsection{TDD in software development} The principal introduction of TDD is described in \cite{Beck04}: it is essentially a methodology of software development, where one writes new code only if an automated test has failed, and  eliminating duplication in doing so. The test come from some place: a feature requirement should have a test specification to go with it, or, more concretely,``If you can't write test for what you are about to code, then you shouldn't even be thinking about coding.'' \cite{Kumar13}. That is, TDD permeates the whole of what was traditionally the waterfall or iterative methodology. Shrivastava and Jain \cite{Shrivastava10} summed up the sequence as follows:  
1) Write a test for a piece of functionality (that was based on a requirement),
2) Run all tests to check that the new test fails,
3) Write relevant code that passes the test,
4) Run the specific test to verify it passes,
5) Refactor the code, and
6) Run all tests to verify that the changes to the code did not change the external behaviour of the software (regression testing).
The important difference with unit tests, is that TDD is a {\em test-first} approach rather than the more commonly known `test-last' approach (design, code, test), and therewith moves into the realm of a methodology of its own permeating all steps in the development rather than only the testing phase with its unit tests for individual classes or components. 

TDD is said to result in being more focussed, improved communication, improved understanding of required software behaviour, and reduced design complexity \cite{Kumar13}. Quantitatively, TDD produced code passes more externally defined tests---i.e, better software quality---and involves less  time spent on debugging, and experiments with students showed that the test-first group wrote more tests and were significantly more productive than the test-last group \cite{Janzen05}.

While all this focuses on the actual programming, the underlying ideas have been applied to conceptual data modelling \cite{Tort10,Tort11}, resulting in a TDD authoring process, compared to, among others, Halpin's conceptual schema design procedure \cite{Halpin01}. Tort and Oliv\'e reworked the test specification into a test-driven way of modelling, where each UML Class Diagram language feature has its own test specification in OCL that involves creating the objects that should, or ought not to, instantiate the UML classes and associations \cite{Tort11}. Also here, the test is supposed to fail first, then the model is updated, and then the test ought to pass. The tool that implements it was evaluated with modellers, which made clear, among others, that more time was spent on developing and revising the conceptual model, in the sense of fixing errors, than on writing the test cases \cite{Tort11}.


\subsubsection{Tests in ontology engineering}

An early explorative work on borrowing the notion of testing from software engineering to apply it to ontology engineering is described in \cite{Vrandecic06}, which explores several adaptation options: testing with the axiom and its negation, formalising CQs, checks by means on integrity constraints, autoepistemic operators, and domain and range assertions. Working with CQs has shown to be, relatively, the most popular approach in the years since. A substantial step in the direction of test-driven ontology engineering was proposed by Ren et al \cite{Ren14}, who analyse CQs for use with SPARQL queries that then would be tested against the ontology. It focuses on finding patterns in the natural language-based CQs, the sample SPARQL queries are querying for individuals only, and the formalisation stops at {\em what} has to be tested, not {\em how} that can, or should, be done. 
Earlier work on CQs and queries include the OntologyTest tool, which allows the user to specify tests to check the functional requirements of the ontology based on the CQs, using ABox instances and SPARQL queries \cite{GarcaRamos09}. Unlike extensive CQ and query patters, it specifies different types of tests focussed on the ABox rather than knowledge represented in the TBox, such as ``instantiation tests'' (instance checking) and ``recovering tests'' (query for a class' individuals) and using mock individuals where applicable \cite{GarcaRamos09}; other instance-oriented test approaches have been proposed as well, although focussing on RDF/Linked Data rather than the OWL ABox \cite{Kontokostas14}. 
 A NeON plugin with similar functionality and methodical steps within the eXtreme Design approach also has been proposed \cite{Blomqvist12}, but not the types of tests. A more basic variant is incorporated in the EFO Validator\footnote{\url{http://www.ebi.ac.uk/fgpt/sw/efovalidator/index.html}} for the Experimental Factor Ontology (EFO), which has tests for presence of a specific class or relation. 
 Neither are based on the principle of TDD where, according to its specification, a test first has to fail, then the code is modified, and then the TDD test should pass. 
 Warrender and Lord's approach \cite{Warrender15} does take that in consideration. They focus on unit tests for TBox testing where each query to the ontology requires a new test declaration. The tests have to be specified in Clojure with its own unfamiliar Tawny-Owl notation, describes only subsumption tests although the Karyotype ontology it is applied to is in $\mathcal{ALCHI}$, and the tests themselves are very tailored to the actual ontology rather than having reusable `templates' for the tests covering all OWL language features. On the positive side, it can avail of some existing infrastructure for software testing rather than reinventing that technological wheel.

The ontology unit test notion of axiom and its negation of \cite{Vrandecic06} has been used, in a limited sense, in {\em advocatus diaboli}, where in the absence of a disjointness axiom between classes, it shows  the consequences to the modeller to approve or disapprove, and if the latter is selected, the tool adds the disjointness axiom \cite{Ferre12}. Some of the OOPS! pitfalls that the software checks has some test-flavour to it \cite{Poveda12}, such as suggesting possible symmetry. FORZA shows the permissible relations one can add that will not lead to an inconsistency, which is based on domain and range constraints of properties and the selected entities one wants to relate 
\cite{KKG13forza}; that is, also not hypothesising a failure/absence of required knowledge in the TBox as such, though it can be seen as a variant of the domain \& range assertions unit tests in \cite{Vrandecic06}. 

Concerning overarching methodologies, none of the 9 methodologies reviewed by \cite{Garcia10} are TDD-based, nor is the MeltingPoint Garc\'ia et al. propose themselves. A recent Agile-inspired tool and methodology is OntoMaven. Aspect OntoMaven \cite{Paschke15} is an extension to OntoMaven that is based on reusing ideas from Apache Maven, advocated as being both a tool and supporting agile ontology engineering, such as COLM\footnote{\url{http://www.corporate-semantic-web.de/colm.html}}. Regarding tests, besides the OntoMvnTest with `test cases' for the usual syntax checking, consistency, and entailment, the documentation states it should be possible to reuse Maven plug-ins for further test types \cite{Paschke15}, but this has not been followed through yet.
A different Agile-inspired method is, eXtreme Design with content ODPs \cite{Presutti09}, although this is also more a design method for rapid turnaround times rather than test-driven. Likewise, the earlier proposed RapidOWL is based on ``iterative refinement, annotation and structuring of a knowledge base'' \cite{Auer06} rather permeating the test-driven approach throughout the methodology. RapidOWL does mention the notion of ``short releases'', which is very compatible with TDD cf. NeON's waterfall-inspired `write many CQs first' \cite{Suarez08}, but not how this is to be achieved other than publishing new versions quickly.

Thus, full TDD ontology engineering as such has not been proposed yet, to the best of our knowledge. While the idea of unit tests---which potentially could become part of TDD test---has been proposed, there is a dearth of actual specifications as to what exactly is, or should be, going on in such as test. It is also unclear whether even when one were to specify basic tests for each language feature, whether they can be put together in a modular fashion for the more complex axioms that can be declared with OWL 2. Further, there is no regression testing to check that perhaps an earlier modelled CQ---and thus a passed test---conflicts with a later one, and identifying which ones are conflicting.

\section{TDD specification for ontologies}
\label{sec:core}

Before introducing the TBox and RBox TDD tests, first the general procedure in introduced and some clarifications are given on notation, the notion of true/false of a TDD test, and mock entities.

\subsection{Preliminaries}

Taking the TDD approach of devising a test that demonstrates absence of the feature (i.e., test failure), add feature, test again whether the test passes, the generalised TDD principle for ontologies then becomes:
\begin{compactenum}
\item input: CQ and transform this into an axiom (optionally)
\item given: axiom $x$ of type $X$ to be added to the ontology.
\item check the vocabulary elements of $x$ are in ontology $O$ (itself a TDD test)
\item run TDD test twice: 
	\begin{compactenum}
		\item the first execution should fail (check 
		 $O \nvDash x$ or not present), 
		\item update the ontology (add $x$), and 
		\item run the test again which then should pass (check that $O \models x$) and such that there is no new inconsistency or undesirable deduction
	\end{compactenum}
\item Run all previous successful tests, which still have to pass (i.e., regression testing)
\end{compactenum}

In the following two subsections, we define such TDD test for TBox and RBox axioms with respect to OWL 2 DL features. For the TDD tests, there are principally two options: a TDD test at the TBox-level (where possible), or always using individuals explicitly asserted in the ABox. We specify tests for both approaches. For the test specifications, we use the OWL 2 standard notation for the ontology's vocabulary: $C,D,E, ... \in V_{C}$, $R,S, ... \in V_{OP}$, and $a, b,... \in V_I$. 

For the TBox tests, we use SPARQL-OWL \cite{Kollia11} where possible/applicable\footnote{While SPARQL-DL \cite{Sirin07} might be more well-known than SPARQL-OWL, that version does not permit TBox queries with object properties, whereas SPARQL-OWL does, and we need those features}. Its notation is principally reusing OWL functional syntax-style notation merged with SPARQL's queried objects (i.e., {\sf ?x}) for the formulation of the query, and adding a variable for the query answer; e.g., $\alpha \leftarrow$ {\sf SubClassOf (?x D)} will return all subclasses of class {\sf D}. The query rewriting of SPARQL-OWL has been described in \cite{Kollia11} and a tool implementing the algorithms, OWL-BGP, is freely available\footnote{\url{https://github.com/iliannakollia/owl-bgp}}. While for some tests, one can use Prot\'eg\'e's intuitive DL query tab to perform the tests, this is possible only for a fraction of the tests, and therefore omitted.

Some TBox and all ABox tests will require additional classes or individuals that only serve the purpose of testing and have to be removed after the test terminates successfully. This resembles the notion of {\em mock objects} in software engineering \cite{Mackinnon01,Kim06}, which we shall transport into the ontology setting. While TDD tests for ontologies do not need stubs for class methods, in some cases a test does need additional elements for it to be testable. Therefore, we do use {\em mock class} for a temporary OWL class, {\em mock individual} for an ABox individual created for the TDD test, and {\em mock axiom} for any auxiliary axiom that has to be added solely for testing purposes; they are to be removed from the ontology after completion of the test. (Possibly this can be taken further, using mock ontology modules, alike the mock databases \cite{Taneja10}, but this is not needed at this stage.)

Steps 3 and 4a in the sequence listed above may give an impression of epistemic queries. It has to be emphasised that there is a fine distinction between 1) checking when an element is in the vocabulary of the TBox of the ontology (in $V_C$ or $V_{OP}$) versus autoepistemic queries, and 2) whether something is {\em logically} true or false versus a {\em test} evaluating to true or false. Proposals for epistemic extensions for OWL exist, which concern instance retrieval and integrity constraints checking \cite{MehdiRG11} that reduces the \textbf{\textsf{K}}-operator to plain OWL queries, or closed world reasoning \cite{Grimm05}. The TDD test focus in step 3, above, is of a distinctly different flavour. We need to know whether an element is present in the ontology, but we do not want to `know' or assert things about individuals (say, that when there is an instance of country, that there must also be an object of capital associated with it, as in, \textbf{\textsf{K}}Country $\sqsubseteq$ \textbf{\textsf{A}}$\exists$hasCapital.\textbf{\textsf{A}}Capital). As such, an epistemic query language is not needed for the TBox-level axioms. In the TDD context, the epistemic-sounding `not asserted in or inferred from the ontology' is to be understood in the context of a {\em TDD test}, like whether an ontology has some class $C$ in its vocabulary, not whether it is `known to exist' in one's open or closed world based on the knowledge represented in the ontology.

\subsection{Test patterns for TBox axioms}

The tests are introduced in sequence, where the primed test names are the ones with explicit individuals. 


\subsubsection{Class subsumption, $T_{cs}$ or $T^{\prime}_{cs}$.}
When the axiom to add is of type $C \sqsubseteq D$, then $O \models \neg (C \sqsubseteq D)$ should be true if it were not present. Logically, this means $O \cup \neg\neg (C \sqsubseteq D)$ should be inconsistent, i.e., $O \cup (\neg C \sqcup D)$. 
Given the current Semantic Web technologies, it is easier to just query the ontology for the superclasses of  $C$ and to ascertain that $D$ is not in query answer $\alpha$ rather than create and/or execute tailor-made tableau algorithms. 
Thus, a test $T$ can be true or false. In SPARQL-OWL notation and quasi algorithm for comprehensiveness, $T_{cs}$ is:
\begin{algorithmic}[1]
\Require Test $T(C \sqsubseteq D)$ \Comment{i.e., $T_{cs}$}
\State $\alpha \leftarrow$ {\sf SubClassOf(?x D)}
\If{$C \notin \alpha$} \Comment{{\small $C \sqsubseteq D$ is neither asserted nor entailed in the ontology}}
	\State {\bf return} $T(C \sqsubseteq D)$ is false
\Else
	\State {\bf return} $T(C \sqsubseteq D)$ is true
\EndIf
\end{algorithmic}
%
%
\noindent 
After adding $C \sqsubseteq D$ to the ontology, the same test is run, which should evaluate to $D \in \alpha$ and therewith $T(C \sqsubseteq D)$ is true. 
%

The TTD test with individuals concerns a check whether an instance of $C$ is also an instance of $D$. That is, for $T^{\prime}_{cs}$ we have:
\begin{algorithmic}[1]
\Require Test $T(C \sqsubseteq D)$ \Comment{i.e., $T^{\prime}_{cs}$}
\State Create a mock object $a$
\State Assert $C(a)$
\State $\alpha \leftarrow$ {\sf Type(?x D)}
\If{$a \notin \alpha$} \Comment{{\small $C \sqsubseteq D$ is neither asserted nor entailed in the ontology}}
	\State {\bf return} $T(C \sqsubseteq D)$ is false
\Else
	\State {\bf return} $T(C \sqsubseteq D)$ is true
\EndIf
\end{algorithmic}
%


\subsubsection{Class disjointness, $T_{cd}$ or $T^{\prime}_{cd}$.}
One can assert the complement, $C \sqsubseteq \neg D$, or disjointness, $C \sqcap D \sqsubseteq \bot$. Let us consider the former first, for a test $T_{cd_c}$. For the test, then $\neg (C \sqsubseteq \neg D)$ should be true, or $T(C \sqsubseteq \neg D)$ false (in the sense of `not be in the ontology'), 
Testing for the latter only does not suffice, however, as there are more cases where $O \nvDash C \sqsubseteq D$ holds, but disjointness is not really applicable---being classes in distinct sub-trees in the TBox---or holds when disjointness is asserted already, which is when $C$ and $D$ are sibling classes. For this complement with the inclusion axiom in OWL, we simply can query for the complement in the ontology:
%
\begin{algorithmic}[1]
\Require Test $T(C \sqsubseteq \neg D)$ \Comment{i.e.,  test $T_{cd_c}$}
\State $\alpha \leftarrow$ {\sf ObjectComplementOf(C ?x)}
\If{$D \notin \alpha$} \Comment{thus $O \nvDash C \sqsubseteq \neg D$}
	\State {\bf return} $T(C \sqsubseteq \neg D)$ is false
\Else
	\State {\bf return} $T(C \sqsubseteq \neg D)$ is true	
\EndIf
\end{algorithmic}

\noindent Concerning the stronger version of disjointness, $C \sqcap D \sqsubseteq \bot$, in SPARQL-OWL notation: 
\begin{algorithmic}[1]
\Require Test $T(C \sqcap D \sqsubseteq \bot)$ \Comment{i.e.,  test $T_{cd_d}$}
\State $\alpha \leftarrow$ {\sf DisjointClasses(?x D)} 
\If{$C \notin \alpha$} \Comment{thus, $O \nvDash C \sqcap D \sqsubseteq \bot$}
	\State {\bf return} $T(C \sqcap D \sqsubseteq \bot)$ is false
\Else
	\State {\bf return} $T(C \sqcap D \sqsubseteq \bot)$ is true	
\EndIf
\end{algorithmic}
%
%
The second option for the test, $T^{\prime}_{cd}$, is to involve the ABox and use a query or run the reasoner. 
The sequence of steps is as follows, availing of the reasoner only and no additional queries are necessary:
\begin{algorithmic}[1]
\Require Test $T(C \sqsubseteq \neg D)$ or $T(C \sqcap D \sqsubseteq \bot)$ \Comment{i.e., test $T^{\prime}_{cd}$}
\State Create individual $a$ \Comment{that is, $a$ is a mock object}
\State Assert $C(a)$ and $D(a)$; 
\State $ostate \leftarrow$ Run the reasoner
\If{$ostate$ == consistent} \Comment{{\small the test fails, i.e., then either $O \nvDash C \sqsubseteq \neg D$ or $O \nvDash C \sqcap D \sqsubseteq \bot$ directly or through one or both of their superclasses}} 
	\State {\bf return} $T(C \sqsubseteq \neg D)$ or $T(C \sqcap D \sqsubseteq \bot)$ is false
\Else
	\Comment{{\small the ontology is inconsistent, the test passed; thus either $C \sqsubseteq \neg D$ or $C \sqcap D \sqsubseteq \bot$ is already asserted among both their superclasses or among $C$ or $D$ and a superclass of $D$ or $C$, respectively.}}
	\State {\bf return} $T(C \sqsubseteq \neg D)$ or $T(C \sqcap D \sqsubseteq \bot)$ is true
\EndIf
\end{algorithmic}
%
Further, from a modelling viewpoint, it would make sense to also require $C$ and $D$ to be siblings. The sibling requirement can be added as an extra check in the interface to alert the modeller to it, but not be enforced from a logic viewpoint.


\subsubsection{Class equivalence, $T_{ce}$ and $T^{\prime}_{ce}$.}
When the axiom to add is of the form $C \equiv D$, then $O \models \neg (C \equiv D)$ should be true before the edit, or $O \nvDash C \equiv D$ false. The latter is easier to test, for we can simply run aforementioned 
$T_{cs}$ test twice, once for $C \sqsubseteq D$ and once for $D \sqsubseteq C$: if both are true, then $O \models C \equiv D$, if one of them or neither is true, then $O \nvDash C \equiv D$. 
More succinctly though, one can use the following SPARQL-OWL query: 
\begin{algorithmic}[1]
\Require Test $T(C \equiv D)$ \Comment{i.e.,  test $T_{ce}$}
\State $\alpha \leftarrow$ {\sf EquivalentClasses(?x D)} 
\If{$C \notin \alpha$} \Comment{thus, $O \nvDash C \equiv D$}
	\State {\bf return} $T(C \equiv D)$ is false
\Else
	\State {\bf return} $T(C \equiv D)$ is true	
\EndIf
\end{algorithmic}
%

\noindent Subsequently, it has to be added to the ontology, queried again, and evaluate to $D \in \alpha$.
However, this works only when $D$ is an atomic class, not when $D$ is a complex one. 

For $T^{\prime}_{ce}$ with individuals, we specify an extended version of $T^{\prime}_{cs}$ as follows:
\begin{algorithmic}[1]
\Require Test $T(C \equiv D)$ \Comment{i.e., test $T^{\prime}_{ce}$}
\State Create a mock object $a$
\State Assert $C(a)$
\State $\alpha \leftarrow$ {\sf Type(?x D)}
\If{$a \notin \alpha$} \Comment{thus, $O \nvDash C \equiv D$}
	\State Delete $C(a)$ and $a$
	\State {\bf return} $T(C \equiv D)$ is false
\Else
	\State Delete $C(a)$
	\State Assert $D(a)$
	\State $\alpha \leftarrow$ {\sf Type(?x C)}
	\If{$a \notin \alpha$}   \Comment{thus, $O \nvDash C \equiv D$}
		\State {\bf return} $T(C \equiv D)$ is false
	\Else
		\State {\bf return} $T(C \equiv D)$ is true
	\EndIf
	\State Delete $D(a)$ and $a$
\EndIf
\end{algorithmic}



\subsubsection{Simple existential quantification, $T_{eq}$ or $T^{\prime}_{eq}$.}

Let the axiom type $X$ be of the form $C \sqsubseteq \exists R.D$, then $O \nvDash \neg(C \sqsubseteq \exists R.D)$ should be true, or $O \models C \sqsubseteq \exists R.D$ false (or: not asserted) before the ontology edit. One could do a first check that $D$ is not a descendant of $R$ but if it is, then it may be the case that $C^{\prime} \sqsubseteq \exists R.D$, with $C$ a different class from $C^{\prime}$. This still requires one to confirm that $C$ is not a subclass of $\exists R.D$. In SPARQL-OWL, we can combine this into one query/TDD test:
\begin{algorithmic}[1]
\Require Test $T(C \sqsubseteq \exists R.D)$ \Comment{i.e.,  test $T_{eq}$}
\State $\alpha \leftarrow$ {\sf SubClassOf(?x ObjectSomeValuesFrom(R D))} 
\If{$C \notin \alpha$} \Comment{thus, $O \nvDash C \sqsubseteq \exists R.D$}
	\State {\bf return} $T(C \sqsubseteq \exists R.D)$ is false
\Else
	\State {\bf return} $T(C \sqsubseteq \exists R.D)$ is true	
\EndIf
\end{algorithmic}
%
%
If the test passes, i.e., $C \notin \alpha$, then the axiom is to be added to the ontology, the query run again, and if $C \in \alpha$, then the test cycle is completed.

From a cognitive, or modelling, viewpoint, desiring to add a CQ that amounts to $C \sqsubseteq \exists R.\neg D$ (`each $C$ has an outgoing arc $R$ to anything that is not a $D$') may look different, but $\neg D \equiv D^{\prime}$, so it amounts to testing $C \sqsubseteq \exists R.D^{\prime}$, i.e., essentially the same pattern. This also can be formulated directly into a SPARQL-OWL query, encapsulated in a TDD test: 
\begin{algorithmic}[1]
\Require Test $T(C \sqsubseteq \exists R.\neg D)$ \Comment{i.e.,  test $T_{eq_{nd}}$}
\State $\alpha \leftarrow$ {\sf SubClassOf(?x ObjectSomeValuesFrom(R  ObjectComplementOf(D)))} 
\If{$C \notin \alpha$} \Comment{thus, $O \nvDash C \sqsubseteq \exists R.\neg D$}
	\State {\bf return} $T(C \sqsubseteq \exists R.\neg D)$ is false
\Else
	\State {\bf return} $T(C \sqsubseteq \exists R.\neg D)$ is true	
\EndIf
\end{algorithmic}
%


\noindent It is slightly different for $C \sqsubseteq \neg\exists R.D$ (`all $C$s do not have an outgoing arc $R$ to $D$'). The query with TDD test is as follows: 

\begin{algorithmic}[1]
\Require Test $T(C \sqsubseteq \neg \exists R.D)$ \Comment{i.e.,  test $T_{eq_{nr}}$}
\State $\alpha \leftarrow$ {\sf SubClassOf(?x ObjectComplementOf(ObjectSomeValuesFrom(R  D)))} 
\If{$C \notin \alpha$} \Comment{thus, $O \nvDash C \sqsubseteq \neg \exists R.D$}
	\State {\bf return} $T(C \sqsubseteq \neg \exists R.D)$ is false
\Else
	\State {\bf return} $T(C \sqsubseteq \neg \exists  R.D)$ is true	
\EndIf
\end{algorithmic}
%
%
\noindent hence, then the test fails, so it can be added, the test runs gain, and then passes.
\noindent The TDD test $T^{\prime}_{eq}$ with individuals can be carried out as follows.
\begin{algorithmic}[1]
\Require Test $T(C \sqsubseteq \exists R.D)$ \Comment{i.e., test $T^{\prime}_{eq}$}
\State Create two mock objects, $a$ and $b$
\State Assert $C(a)$, $D(b)$, and $R(a,b)$
\State $\alpha \leftarrow$ {\sf Type(?x, ObjectSomeValuesFrom(R D))}
\If{$a \notin \alpha$}  \Comment{thus, $O \nvDash C \sqsubseteq \exists R.D$}
	\State {\bf return} T($C \sqsubseteq \exists R.D$) is false
\Else
	\State {\bf return} T($C \sqsubseteq \exists R.D$) is true
\EndIf	
\State Delete $C(a)$, $D(b)$, $R(a,b)$, $a$, and $b$
\end{algorithmic}
The two negated cases are as follows.
\begin{algorithmic}[1]
\Require Test $T(C \sqsubseteq \exists R.\neg D)$ \Comment{i.e., test $T^{\prime}_{eq_{nd}}$}
\State Create two mock objects, $a$ and $b$
\State Assert $C(a)$, $\neg D(b)$, and $R(a,b)$
\State $\alpha \leftarrow$ {\sf Type(?x,  ObjectSomeValuesFrom(R ObjectComplementOf(D)))} 
\If{$a \notin \alpha$}  \Comment{thus, $O \nvDash C \sqsubseteq \exists R.\neg D$}
	\State {\bf return} T($C \sqsubseteq \exists R.\neg D$) is false 
\Else
	\State {\bf return} T($C \sqsubseteq \exists R.\neg D$) is true 
\EndIf	
\State Delete $C(a)$, $D(b)$, $R(a,b)$, $a$, and $b$
\end{algorithmic}
\begin{algorithmic}[1]
\Require Test $T(C \sqsubseteq \neg \exists R.D)$ \Comment{i.e., test $T^{\prime}_{eq_{nr}}$}
\State Create two mock objects, $a$ and $b$
\State Assert $C(a)$, $D(b)$, and $R(a,b)$
\State $ostate \leftarrow$ Run the reasoner
\If{$ostate ==$ consistent}  \Comment{thus, $O \nvDash C \sqsubseteq \neg \exists R.D$}
	\State {\bf return} T($C \sqsubseteq \neg\exists R.D$) is false
\Else
	\State {\bf return} T($C \sqsubseteq \neg\exists R.D$) is true
\EndIf	
\State Delete $C(a)$, $D(b)$, $R(a,b)$, $a$, and $b$
\end{algorithmic}


\subsubsection{Simple universal quantification, $T_{uq}$ or $T^{\prime}_{uq}$.}
Let $X$ be $C \sqsubseteq \forall R.D$, then $O \nvDash \neg(C \sqsubseteq \forall R.D)$ should hold, or $O \models C \sqsubseteq \forall R.D$ false (not be present in the ontology), before the ontology edit. This has a similar pattern to the one for existential quantification, 
\begin{algorithmic}[1]
\Require Test $T(C \sqsubseteq \forall R.D)$ \Comment{i.e.,  test $T_{uq}$}
\State $\alpha \leftarrow$ {\sf SubClassOf(?x ObjectAllValuesFrom(R D))} 
\If{$C \notin \alpha$} \Comment{thus, $O \nvDash C \sqsubseteq \forall R.D$}
	\State {\bf return} $T(C \sqsubseteq \forall R.D)$ is false
\Else
	\State {\bf return} $T(C \sqsubseteq \forall R.D)$ is true	
\EndIf
\end{algorithmic}
%
%
\noindent which then can be added and the test ran again. 
 %
The TDD test for $T^{\prime}_{uq}$ with individuals is much alike $T^{\prime}_{eq}$: 
\begin{algorithmic}[1]
\Require Test $T(C \sqsubseteq \forall R.D)$ \Comment{i.e., test $T^{\prime}_{uq}$}
\State Create two mock objects, $a$ and $b$\
\State Assert $C(a)$, $D(b)$, and $R(a,b)$
\State $\alpha \leftarrow$ {\sf Type(?x,  ObjectAllValuesFrom(R D))}
\If{$a \notin \alpha$}  \Comment{thus, $O \nvDash C \sqsubseteq \forall R.D$}
	\State {\bf return} T($C \sqsubseteq \forall R.D$) is false
\Else
	\State {\bf return} T($C \sqsubseteq \forall R.D$) is true
\EndIf	
\State Delete $C(a)$, $D(b)$, $R(a,b)$, $a$, and $b$
\end{algorithmic}


\subsection{Test patterns for object propery axioms}

From here onward, the tests deal with object properties more specifically---sometimes called the RBox---that more often than not do not lend themselves well for easy querying in the DL query tab, for the DL query tab returns results on classes and individuals only, because DL query is essentially a class expression language, i.e., it can only express (complex) classes. However, SPARQL-OWL and the automated reasoner {\em can} be used fairly straightforwardly, so it is essentially an interface limitation that can be solved with adding a new plugin for TDD. 


\subsubsection{Domain axiom, $T_{da}$ or $T^{\prime}_{da}$.} Let $X$ be $\exists R \sqsubseteq C$ that is not yet in $O$. To verify that it is not, $O \models \neg(\exists R \sqsubseteq C)$ should be true, or $O \models  \exists R \sqsubseteq C$ false. With SPARQL-OWL, there are two options. First, one can query for the domain:
\begin{algorithmic}[1]
\Require Test $T(\exists R \sqsubseteq C)$ \Comment{i.e.,  test $T_{da}$}
\State $\alpha \leftarrow$ {\sf ObjectPropertyDomain(R ?x)} 
\If{$C \notin \alpha$} \Comment{thus, $O \nvDash \exists R \sqsubseteq C$}
	\State {\bf return} $T(\exists R \sqsubseteq C)$ is false
\Else
	\State {\bf return} $T(\exists R \sqsubseteq C)$ is true	
\EndIf
\end{algorithmic}
%
%
Alternatively, one can query for the superclasses of $\exists R$ (noting that it is shorthand for $\exists R.\top$), where the above-listed query in the TDD is replaced with:\\
%
	\indent  $\alpha \leftarrow$ {\sf SubClassOf(SomeValuesFrom(R Thing) ?x)}\\
%
Note that  $C \in \alpha$ only will be returned if $C$ is the only domain class of $R$ or when $C \sqcap C^{\prime}$ (but not if it is $C \sqcup C^{\prime}$, which is a superclass of $C$). 
%
%
Alternatively, with the individuals:
\begin{algorithmic}[1]
\Require Test $T(\exists R \sqsubseteq C)$ \Comment{i.e., test $T^{\prime}_{da}$}
\State Check $R \in V_{OP}$ and $C \in V_{C}$
\State Add individuals $a$ and $topObj$
\State Add $R(a,topObj)$ as object property assertion
\State Run the reasoner
\If{$a \notin C$} \Comment{{\small $O \nvDash \exists R \sqsubseteq C$ (also in the strict sense as is or with a conjunction), hence then the test fails as intended.}}
	\State {return} $T(\exists R \sqsubseteq C)$ is false
\Else
	\State {return} $T(\exists R \sqsubseteq C)$ is true
\EndIf
\State Delete individuals $a$ and $topObj$
\end{algorithmic}
%
If the answer is empty, then $R$ does not have any domain specified yet, and if $C \notin \alpha$, then $O \nvDash  \exists R \sqsubseteq C$, so that it can be added and the test run again to complete the cycle. 
 
 
\subsubsection{Range axiom, $T_{ra}$ or $T^{\prime}_{ra}$.} When $X$ is a range axiom, $\exists R^- \sqsubseteq D$ should not be in the ontology before the TDD test. This is similar to the domain axiom test:
\begin{algorithmic}[1]
\Require Test $T(\exists R^- \sqsubseteq D)$ \Comment{i.e.,  test $T_{ra}$}
\State $\alpha \leftarrow$ {\sf ObjectPropertyRange(R ?x)} 
\If{$D \notin \alpha$} \Comment{thus, $O \nvDash \exists R^- \sqsubseteq D$}
	\State {\bf return} $T(\exists R^- \sqsubseteq D)$ is false
\Else
	\State {\bf return} $T(\exists R^- \sqsubseteq D)$ is true	
\EndIf
\end{algorithmic}
%
%
or, in the second option, to replace the TDD query with\\
	\indent  $\alpha \leftarrow$ {\sf SubClassOf(SomeValuesFrom(ObjectInverseOf(R) Thing) ?x)}\\
%
The answer will be $D \in \alpha$ if $O \models \exists R^- \sqsubseteq D$ or $O \models \exists R^- \sqsubseteq D \sqcap D^{\prime}$, and only {\tt owl:Thing} $\in \alpha$ if no range was declared for $R$. 
%
%
The test with individuals is as follows:
\begin{algorithmic}[1]
\Require Test $T(\exists R^- \sqsubseteq D)$ \Comment{i.e., test $T^{\prime}_{ra}$}
\State Check $R \in V_{OP}$ and $D \in V_{C}$
\State Add individual $a$ and $topObj$ to the ABox
\State Add $R(topObj,a)$ as object property assertion
\If{$a \notin D$} \Comment{$O \nvDash \exists R^- \sqsubseteq D$, as intended}
	\State {return} $T(\exists R^- \sqsubseteq D)$ is false
\Else
	\State {return} $T(\exists R^- \sqsubseteq D)$ is true
\EndIf
\State Delete $R(topObj,a)$ and individuals $a$ and $topObj$
\end{algorithmic}


\subsubsection{Object property subsumption and equivalence, $T_{ps}$ and $T_{pe}$, and $T^{\prime}_{ps}$ and $T^{\prime}_{pe}$.}
When axiom type $X$ is a property subsumption, $R \sqsubseteq S$, then we have to test that $O \models \neg(R \sqsubseteq S)$, or that $R \sqsubseteq S$ fails. The SPARQL-OWL query in the $T_{ps}(R \sqsubseteq S)$ is as follows:
\begin{algorithmic}[1]
\Require Test $T(R \sqsubseteq S)$ \Comment{i.e.,  test $T_{ps}$}
\State $\alpha \leftarrow$ {\sf SubObjectPropertyOf(?x S)} 
\If{$R \notin \alpha$} \Comment{thus, $O \nvDash R \sqsubseteq S$}
	\State {\bf return} $T(R \sqsubseteq S)$ is false
\Else
	\State {\bf return} $T(R \sqsubseteq S)$ is true	
\EndIf
\end{algorithmic}
%
%

One cannot query this as such in Prot\'eg\'e's DL query tab, other than by using individuals, i.e., a version of $T^{\prime}_{ps}$. 
With the OWL semantics, for $R \sqsubseteq S$ to hold, it means that, given some individuals $a$ and $b$, that if $R(a,b)$ then $S(a,b)$. 
The desired result is computed by the reasoner anyhow. The steps for the TDD test: 
\begin{algorithmic}[1]
\Require Test $T(R \sqsubseteq S)$ \Comment{i.e., test $T^{\prime}_{ps}$}
\State Check $R,S \in V_{OP}$
\State Add individuals $a,b$ to the ABox, add $R(a,b)$
\State Run the reasoner
\If{$S(a,b) \notin \alpha$} \Comment{{\small (practically: not shown in the ``property assertions'' in the individuals tab for $a$, with ``Show inferences'' checked), thus $O \nvDash R \sqsubseteq S$}}
	\State {\bf return} $T(R \sqsubseteq S)$ is false
\Else
	\State {\bf return} $T(R \sqsubseteq S)$ is true
\EndIf
\State Delete $R(a,b)$, and individuals $a$ and $b$
\end{algorithmic}
%
Then the modeller would add $R \sqsubseteq S$, run the test again, and it should then infer $S(a,b)$, as $O \models R \sqsubseteq S$. This, however, does not guarantee $R \sqsubseteq S$ was added, and not inadvertently $R \equiv S$. Their difference can be easily observed with the following set-up:
\begin{algorithmic}[1]
\Require Test $T(R \equiv S)$ \Comment{i.e., test $T^{\prime}_{pe}$}
\State Check $R,S \in V_{OP}$
\State Add mock individuals $a,b,c,d$ to the ABox
\State Add $R(a,b)$ and $S(c,d)$ as object property assertion
\State Run the reasoner
\If{$S(a,b) \in \alpha$ {\bf and} $R(c,d) \notin \alpha$} \Comment{{\small $O \models R \sqsubseteq S$, hence the ontology edit executed correctly}}
	\State {\bf return} $T(R \equiv S)$ is false
\Else \Comment{i.e. $\{S(a,b), R(c,d)\} \in \alpha$, so $O \models R \equiv S$}
	\State $T(R \equiv S)$ is true
\EndIf	
\State Delete $R(a,b)$ and $S(c,d)$, and $a,b,c,d$
\end{algorithmic}

\noindent For object property equivalence at the Tbox level, i.e., $R \equiv S$, one could use $T_{ps}$ twice, or simply use the EquivalentObjectProperties with SPARQL-OWL:  
\begin{algorithmic}[1]
\Require Test $T(R \equiv S)$ \Comment{i.e.,  test $T_{pe}$}
\State $\alpha \leftarrow$ {\sf EquivalentObjectProperties(?x S)} 
\If{$R \notin \alpha$} \Comment{thus, $O \nvDash R \equiv S$}
	\State {\bf return} $T(R \equiv S)$ is false
\Else
	\State {\bf return} $T(R \equiv S)$ is true	
\EndIf
\end{algorithmic}

\subsubsection{Object property inverses, $T_{pi}$ and $T^{\prime}_{pi}$.} There are two options here since OWL 2, and it is hard to choose which one is `better': using explicit inverses tends to be chosen for understanding (e.g., {\tt teaches} with inverse declared explicitly as {\tt taught by}), whereas using an `implicit' inverse (e.g., {\tt teaches} and {\tt teaches$^-$}) improved reasoner performance in at least one instance \cite{Keet14ore}. Practically, for the failure-test of TDD, we can test only the former case, as the latter is only used in axioms. 
Also here one can choose between a TBox or an ABox approach. The TBox approach with a SPARQL-OWL query for $T_{pi}(R \sqsubseteq S^-)$:
\begin{algorithmic}[1]
\Require Test $T(R \sqsubseteq S^-)$ \Comment{i.e.,  test $T_{pi}$}
\State $\alpha \leftarrow$ {\sf InverseObjectProperties(?x S)} 
\If{$R \notin \alpha$} \Comment{thus, $O \nvDash R \sqsubseteq S^-$}
	\State {\bf return} $T(R \sqsubseteq S^-)$ is false
\Else
	\State {\bf return} $T(R \sqsubseteq S^-)$ is true	
\EndIf
\end{algorithmic}
%
Using the Abox, we again have to work with mock objects:
\begin{algorithmic}[1]
\Require test $T(R \sqsubseteq S^-)$ \Comment{i.e., test $T^{\prime}_{pi}$}
\State Check $R,S \in V_{OP}$
\State \Comment{{\small Assume $S$ is intended to be the inverse of $R$ (with $R$ and $S$ having different names), and we check for its absence:}}
\State Add mock individuals $a,b$ to the ABox
\State Add $R(a,b)$ as object property assertion
\State Run the reasoner
\If{$O \nvDash S(b,a)$} \Comment{$O \nvDash R \sqsubseteq S^-$, hence the test fails, as intended}
	\State {\bf return} $T(R \sqsubseteq S^-)$ is false
\Else
	\State {\bf return} $T(R \sqsubseteq S^-)$ is true
\EndIf
\State Delete mock individuals $a$ and $b$	
\end{algorithmic}
Then add $R \sqsubseteq S^-$, run the test again, which then should evaluate to true.

\subsubsection{Object property chain, $T_{pc}$ or $T^{\prime}_{pc}$.}
When $\mathcal{X}$ is one of the permissible chains (except for transitivity; see below), such as $R \circ S \sqsubseteq S$, $S \circ R \sqsubseteq S$, $R \circ S_1 \circ ... \circ S_n \sqsubseteq S$ (with $n>1$). 
This is increasingly more cumbersome to test, for the simple fact that many more entities are involved, hence, more opportunity to have incomplete knowledge represented in the ontology and, hence, more hassle to find all the permutations that lead to not having the desired effect. Perhaps the easiest way to check whether the ontology has the property chain, is to search the owl file for {\tt owl:propertyChainAxiom}, with the relevant properties included in order, or, for that matter, the SPARQL-OWL query in the TDD test:
\begin{algorithmic}[1]
\Require Test $T(R \circ S \sqsubseteq S)$ \Comment{i.e.,  test $T_{pc}$}
\State $\alpha \leftarrow$ {\sf SubObjectPropertyOf(ObjectPropertyChain(R S) ?x)} 
\If{$S \notin \alpha$} \Comment{thus, $O \nvDash R \circ S \sqsubseteq S$}
	\State {\bf return} $T(R \circ S \sqsubseteq S)$ is false
\Else
	\State {\bf return} $T(R \circ S \sqsubseteq S)$ is true	
\EndIf
\end{algorithmic}
%
%
and similarly with the other permutations.  However, simply checking the owl file or executing the query misses three aspects of chains: i) there is no point in having a property chain if the properties involved are never used in the intended way anyway, 2) this cannot ascertain that it does only what was intended, and 3) whether the chain is allowed in OWL 2, i.e., not violating `interfering' constraints (the properties have to be `simple'). 
For $O \models R \circ S \sqsubseteq S$ to be interesting for the ontology, also at least one $O \models C \sqsubseteq \exists R.D$ and one $O \models D \sqsubseteq \exists S.E$ should be present. If they all were, then a SPARQL-OWL query\\
\indent $\alpha \leftarrow$ {\sf SubClassOf(?x ObjectSomeValuesFrom(S E))}\\
%
will have $C \in \alpha$. 
If either of the three axioms are not present, then $C \notin \alpha$. 
From the perspective of the ABox testing, we need the following sequence of steps:
\begin{algorithmic}[1]
\Require Test $T(R \circ S \sqsubseteq S)$ \Comment{i.e., test $T^{\prime}_{pc}$}
\State Check $R,S \in V_{OP}$ and $C,D,E \in V_C$
\If{$C,D, E \notin V_C$}
	\State Add the missing class(es) ($C$, $D$, and/or $E$) as mock classes 
\EndIf
\State Run the test $T_{eq}$ or $T^{\prime}_{eq}$, for both $C \sqsubseteq \exists R.D$ and for $D \sqsubseteq \exists S.E$
\If{$T_{eq}$ is false}
	\State Add $C \sqsubseteq \exists R.D$, $D \sqsubseteq \exists S.E$, or both, as mock axiom
\EndIf
\If{$O \models C \sqsubseteq \exists S.D$} \Comment{{\small then test is meaningless, for it would not test the property chain}}
	\State Add mock class $C^{\prime}$, mock axiom $C^{\prime} \sqsubseteq \exists R.D$	
	\State Verify with $T_{eq}$ or $T^{\prime}_{eq}$
	\State $\alpha \leftarrow$ {\sf SubClassOf(?x ObjectSomeValuesFrom(S E))}
	\If{$C^{\prime} \notin \alpha$} \Comment{thus, $O \nvDash R \circ S \sqsubseteq S$}
		\State {\bf return} $T(R \circ S \sqsubseteq S)$ is false
	\Else
		\State {\bf return} $T(R \circ S \sqsubseteq S)$ is true
	\EndIf
\Else \Comment{so, $O \nvDash C \sqsubseteq \exists S.D$}  
	\State $\alpha \leftarrow$ {\sf SubClassOf(?x ObjectSomeValuesFrom(S E))}
	\If{$C \notin \alpha$} \Comment{thus, $O \nvDash R \circ S \sqsubseteq S$}
		\State {\bf return} $T(R \circ S \sqsubseteq S)$ is false
	\Else 
		\State {\bf return} $T(R \circ S \sqsubseteq S)$ is true
	\EndIf	
\EndIf	
\State Delete all mock objects, classes, and axioms
\end{algorithmic}
Assuming that the test fails, i.e., $C \notin \alpha$ (resp. $C^{\prime} \notin \alpha$) and thus $O \nvDash R \circ S \sqsubseteq S$, then add the property chain and run the test again, which then should pass (i.e., $C \in \alpha$). When it does, any mock classes and axioms should be removed.

The procedure holds similarly for the other permissible combinations of object properties in a property chain/complex role inclusion.

\subsubsection{Object property characteristics, $T_{p_x}$.} Testing absence/presence of the object property characteristics is surely feasible with mock individuals in the ABox, but is doable only for transitivity and local reflexivity in the TBox.

{\em $R$ is functional, $T^{\prime}_{p_f}$}, i.e., some object has at most one individual $R$-successor. The TDD test procedure is as follows. 
\begin{algorithmic}[1]
\Require Test $T(\mbox{\sf Func}(R))$ \Comment{i.e., test $T^{\prime}_{p_f}$}
\State Check $R \in V_{OP}$ and $a,b,c \in V_I$; if not present, add.
\State Assert mock axioms $R(a,b)$, $R(a,c)$, and $b \neq c$, if not present already.
\State Run reasoner
\If{$O$ is consistent} \hfill \Comment{thus, $O \nvDash \mbox{\sf Func}(R)$}
	\State {\bf return} $T(\mbox{\sf Func}(R))$ is false
	\Else \hfill \Comment{$O$ is inconsistent}
		\State {\bf return} $T(\mbox{\sf Func}(R))$ is true
\EndIf
\State Remove mock axioms and individuals, as applicable
\end{algorithmic}


{\em $R$ is inverse functional, $T^{\prime}_{p_{if}}$.} This is as above, but then in the other direction, i.e., $R(b,a)$, $R(c,a)$ with $b,c$ declared distinct, will result in a consistent ontology when $O \nvDash \mbox{\sf InvFun}(R)$. Thus:
\begin{algorithmic}[1]
\Require Test $T(\mbox{\sf InvFun}(R))$ \Comment{i.e., test $T^{\prime}_{p_{if}}$}
\State Check $R \in V_{OP}$ and $a,b,c \in V_I$; if not present, add.
\State Assert mock axioms $R(b,a)$, $R(c,a)$, and $b \neq c$, if not present already.
\State Run reasoner
\If{$O$ is consistent} \hfill \Comment{thus, $O \nvDash \mbox{\sf InvFun}(R)$}
	\State {\bf return} $T(\mbox{\sf InvFun}(R))$ is false
	\Else \hfill \Comment{$O$ is inconsistent}
	\State {\bf return} $T(\mbox{\sf InvFun}(R))$ is true
\EndIf
\State Remove mock axioms and individuals, as applicable
\end{algorithmic}

{\em $R$ is transitive, $T_{p_t}$ or $T^{\prime}_{p_t}$}, so that with $R(a,b)$ and $R(b,c)$, it will infer $R(a,c)$. As with the object property chain test ($T_{pc}$), this object property characteristic is only `interesting' for the ontology engineer if there are at least two related axioms so that one obtains a non-empty deduction thanks to the transitive object property. Further, it is the only object property characteristic that has a real effect in the TBox. If the relevant axioms are not asserted, they have to be added.
\begin{algorithmic}[1]
\Require Test $T(\mbox{\sf Trans}(R))$ \hfill \Comment{i.e., test $T_{p_t}$}
\State Check $R \in V_{OP}$ and $C, D, E,  \in V_C$
\If{$C, D, E,  \notin V_C$}
	\State Add the missing class(es) ($C$, $D$, and/or $E$ as mock classes)
\EndIf
\If{$C \sqsubseteq \exists R.D$ and $D \sqsubseteq \exists R.E$ are not asserted}
	\State add $C \sqsubseteq \exists R.D$ and $D \sqsubseteq \exists R.E$ to $O$
\EndIf
\State $\alpha \leftarrow$ {\sf SubClassOf(?x ObjectSomeValuesFrom(R E))}
\If{$C \notin \alpha$} \hfill \Comment{thus, $O \nvDash \mbox{\sf Trans}(R)$}
	\State {\bf return} $T(\mbox{\sf Trans}(R))$ is false 
	\Else
		\State {\bf return} $T(\mbox{\sf Trans}(R))$ is true
\EndIf		
\State Remove mock classes and axioms, as applicable
\end{algorithmic}
%
The ABox-based test is as follows.
\begin{algorithmic}[1]
\Require Test $T(\mbox{\sf Trans}(R))$ \hfill \Comment{i.e., test $T^{\prime}_{p_t}$}
\State Check $R \in V_{OP}$, $a,b,c \in V_I$. If not, introduce $a,b,c$ as mock objects.
\State Assert mock axioms $R(a,b)$ and $R(b,c)$, if not present already.
\State Run reasoner
\If{$R(a,c) \notin \alpha$} \hfill \Comment{thus, $O \nvDash \mbox{\sf Trans}(R)$}
	\State {\bf return} $T(\mbox{\sf Trans}(R))$ is false 
\Else
	\State {\bf return} $T(\mbox{\sf Trans}(R))$ is true 
\EndIf
\State Remove mock individuals
\end{algorithmic}


{\em $R$ is symmetric, $T^{\prime}_{p_s}$}, so that with $R(a,b)$, it will infer $R(b,a)$. The test-to-fail---assuming $R \in V_{OP}$---is as follows.
\begin{algorithmic}[1]
\Require Test $T(\mbox{\sf Sym}(R))$ \hfill \Comment{i.e., test $T^{\prime}_{p_s}$}
\State Check $R \in V_{OP}$. Introduce $a,b$ as mock objects ($a,b \in V_I$).
\State Assert mock axiom $R(a,b)$.
\State $\alpha \leftarrow$ {\sf ObjectPropertyAssertion(R x? a)}\label{line:qsym} 
\If{$b \notin \alpha$} \hfill \Comment{thus, $O \nvDash \mbox{\sf Sym}(R)$}
	\State {\bf return} $T(\mbox{\sf Sym}(R))$ is false 
\Else
	\State {\bf return} $T(\mbox{\sf Sym}(R))$ is true 
\EndIf
\State Remove mock assertions and individuals
\end{algorithmic}
Alternative to the query in line~\ref{line:qsym}, one can check in the ODE whether $R(b,a)$ is inferred (yellow in the Prot\'eg\'e Individuals tab).

{\em $R$ is asymmetric, $T^{\prime}_{p_a}$}. This is easier to test with the negative, i.e., assert objects symmetrically and distinct, and if the ontology is not inconsistent, then $O \nvDash \mbox{\sf Asym}(R)$. More precisely:
\begin{algorithmic}[1]
\Require Test $T(\mbox{\sf Asym}(R))$ \hfill \Comment{i.e., test $T^{\prime}_{p_a}$}
\State Check $R \in V_{OP}$. Introduce $a,b$ as mock objects ($a,b \in V_I$).
\State Assert mock axioms $R(a,b)$ and $R(b,a)$.
\State Run reasoner
\If{$O$ not inconsistent}\hfill \Comment{thus, $O \nvDash \mbox{\sf Asym}(R)$}
	\State {\bf return} $T(\mbox{\sf Asym}(R))$ is false 
\Else
	\State {\bf return} $T(\mbox{\sf Asym}(R))$ is true 
\EndIf
\State Remove mock axioms and individuals
\end{algorithmic}
%

{\em $R$ is reflexive, $T^{\prime}_{p_{rg}}$ or $T^{\prime}_{p_{rg}}$}. The object property can be either globally reflexive ({\sf Ref($R$)}), or locally ($C \sqsubseteq \exists R.\mbox{Self}$). Global reflexivity is typically not what one wants, but if in the exceptional case the modeller does, then the following test should be executed.
\begin{algorithmic}[1]
\Require Test $T(${\sf Ref($R$)}$)$ \hfill \Comment{i.e., test $T^{\prime}_{p_{rg}}$}
\State Check $R \in V_{OP}$. 
\State Introduce $a$ as mock objects ($a \in V_I$).
\State Run the reasoner
\If{$R(a,a) \notin O$} \hfill \Comment{thus, $O \nvDash {\sf Ref}(R)$}
	\State {\bf return} $T(${\sf Ref($R$)}$)$ is false 
\Else
	\State {\bf return} $T(${\sf Ref($R$)}$)$ is true 
\EndIf
\State Remove mock object $a$
\end{algorithmic}
And adding {\sf Ref($R$)} will have the test evaluate to true. Local reflexivity uses {\sf Self} in an axiom; that is, we need to check whether $O \models C \sqsubseteq \exists R.\mbox{Self}$. This  is essentially the same as $T_{eq}$ but then with {\sf Self} instead of the explicit class, i.e.:
\begin{algorithmic}[1]
\Require Test $T(C \sqsubseteq \exists R.\mbox{Self})$ \Comment{i.e.,  test $T_{p_{rl}}$}
\State $\alpha \leftarrow$ {\sf SubClassOf(?x ObjectSomeValuesFrom(R Self))} 
\If{$C \notin \alpha$} \Comment{thus, $O \nvDash C \sqsubseteq \exists R.\mbox{Self}$}
	\State {\bf return} $T(C \sqsubseteq \exists R.\mbox{Self})$ is false
\Else
	\State {\bf return} $T(C \sqsubseteq \exists R.\mbox{Self})$ is true	
\EndIf
\end{algorithmic}
%
or, in ABox variant:  
\begin{algorithmic}[1]
\Require Test $T(C \sqsubseteq \exists R.\mbox{Self})$ \hfill \Comment{i.e., test $T^{\prime}_{p_{rl}}$}
\State Check $R \in V_{OP}$. Introduce $a$ as mock objects ($a \in V_I$).
\State Assert mock axiom $C(a)$
\State $\alpha \leftarrow$ {\sf Type(?x C), PropertyValue(a R ?x)}
\If{$a \notin \alpha$}\hfill \Comment{thus, $O \nvDash C \sqsubseteq \exists R.\mbox{Self}$}
	\State {\bf return} $T(C \sqsubseteq \exists R.\mbox{Self})$ is false 
\Else
	\State {\bf return} $T(C \sqsubseteq \exists R.\mbox{Self})$ is true 
\EndIf
\State Remove $C(a)$ and mock object $a$
\end{algorithmic}
%

{\em $R$ is irreflexive, $T^{\prime}_{p_{ir}}$}. As with asymmetry, this is easier to test with the converse: assert $R(a,a)$, run the reasoner, then the ontology is consistent (i.e., then $O \nvDash \mbox{\sf Irr}(R)$). Add $\mbox{\sf Irr}(R)$, run the reasoner, then $O \models \bot$, and finally remove mock individual and assertion.
\begin{algorithmic}[1]
\Require Test $T(\mbox{\sf Irr}(R))$ \Comment{i.e., test $T^{\prime}_{p_i}$}
\State Check $R \in V_{OP}$, and add $a \in V_I$
\State Assert mock axiom $R(a,a)$
\State Run reasoner
\If{$O$ is consistent} \hfill \Comment{thus, $O \nvDash \mbox{\sf Irr}(R)$}
	\State {\bf return} $T(\mbox{\sf Irr}(R))$ is false
	\Else \hfill \Comment{$O$ is inconsistent}
		\State {\bf return} $T(\mbox{\sf Irr}(R))$ is true
\EndIf
\State Remove mock axiom and individual, as applicable
\end{algorithmic}


This concludes the basic tests. While the logic permits that some class $C$ on the left-hand side of the inclusion axiom may be  a complex and defined concept, we do not consider such cases here, as due to the tool design of the most widely used ODE, Prot\'eg\'e, the left-hand side of the inclusion axiom has only a single class $C$. 

\section{Experimental evaluation with a Prot\'eg\'e plugin for TDD}
\label{sec:eval}

We describe the design of the plugin and evaluation of the performance tests in this section.

\subsection{Design}

\begin{figure}[t]
	\centering
		\includegraphics[width=1.0\textwidth]{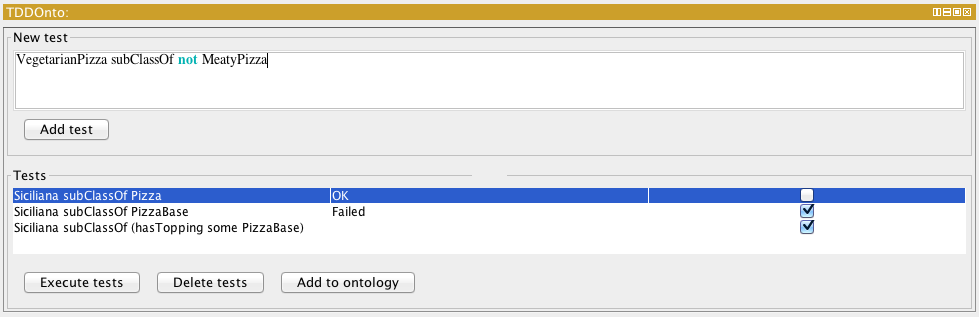}
		\caption{A screenshot of the TDDOnto plugin.}
	\label{fig:plugin}
\end{figure}

In order to support ontology engineers in performing TDD, we have implemented the Prot\'eg\'e plugin named TDDOnto. 
The plugin provides a view where the user may specify the set of tests to be run. 
After their execution, the status of the tests is displayed.
It is also possible to add a selected axiom to the ontology (and re-run the test).
Fig. \ref{fig:plugin} presents the screenshot of the TDDOnto plugin.


\subsection{Evaluation of the TBox and ABox based TDD tests}

The aim of the evaluation is to assess which option is the more feasible one. This can be done from a plugin usability viewpoint and from the algorithmic viewpoint, in the sense of which type of test is faster (though slow response times slow down the ontology authoring process and is also a dimension of usability). We take the latter option here. 
The question, then, is {\em Which TDD approach---queries or mock objects---is better?} We describe the set-up of the experiment first, and then proceed to the results and discussion.

\subsubsection{Set-up of the experiment.}

To assess this quantitatively, we pose the following two  general hypotheses:
\begin{compactenum}
\item[H1:] Query-based TDD is faster overall.
\item[H2:] Classification time of the ontology contributes the most to overall performance (time) of a TDD test.
\end{compactenum}
The reason why we think that H1 will hold is because once classified, one can query multiple times without having to classify the ontology again, and for some mock-object TDD tests the ontology should be inconsistent, which is a more cumbersome step to deal with than checking membership of a class or individual in a query answer. The reason for expecting H2 to hold is that the other operations---adding and removing entities, testing for membership---can be executed in linear time, whereas there are not many ontologies in a language that is linear or less in data complexity. These general hypotheses can be refined to suit statistical tests for each hypotheses:
\begin{compactenum}
\item[H1$_0$:] There is no difference between query-based and mock-object based TDD tests.
\item[H1$_a$:] There is a difference, with query-based having lower values for time taken to execute the tests.
\end{compactenum}
\begin{compactenum}
\item[H2$_0$:] TDD overall execution times are arbitrarily subdivided into ontology classification time and the TDD test part.
\item[H2$_a$:] There is a difference, with ontology classification time taking much more ($>>50\%$) of the TDD overall execution times than the TDD test part.
\end{compactenum}
%

The performance is expected to depend on the ontology's content that is being revised, as reasoning time does.
It is unclear whether 
the overall TDD test execution time and what is attributed to plain ontology classification---also depends on the characteristics of the ontology. 
If it does, it is due to something internal to the reasoner, to which we do not have access. Notwithstanding, we would like to obtain an indication whether there might be interference regarding this aspect. 
Therefore we categorise test ontologies into groups depending on the number of their axioms. 

Testing for/adding the axioms to the selected ontologies can be done in two ways: adding new elements, or reusing elements of the ontology. The former is certainly easier to carry out, the latter is truer to the intended use of a TDD test. In the experiments we followed the latter option and randomly selected existing ontology  classes and properties for the tests.

\subsubsection{Materials and methods.} The results presented in this section are computed based on the tests perfomed on 67 OWL ontologies from the TONES repository \footnote{\url{http://rpc295.cs.man.ac.uk:8080/repository/browser}}, downloaded from the mirror available at OntoHub \footnote{\url{https://ontohub.org/repositories}}. 
We selected those ontologies from all available TONES ontologies, while omitting the other ones that were either not in OWL (but in OBO format) or were having datatypes incompatible with OWL 2, causing exceptions of the reasoner or out of memory exceptions.   
The results were computed on Mac Book Air with 1.3 GHz  Intel Core i5  CPU and 4 GB RAM. 
As an OWL reasoner, we used the same reasoner that is built-in into OWL-BGP, namely HermiT 1.3.8.

The tests were generated randomly, and each test kind was repeated 3 times to obtain more reliable results.
We divided our ontologies into 4 groups, depending on the overall number of their axioms: up to 100 (20 ontologies), 100--1000 axioms (35 ontologies), 1000--10,000 axioms (10 ontologies), and over 10,000 (2 ontologies).
All the experimental results are available at \url{https://semantic.cs.put.poznan.pl/wiki/aristoteles/doku.php}.

\subsubsection{Results and discussion.}

During our experiments we also found out that not all the features of OWL 2 are covered by OWL-BGP.
In particular, RBox axioms (e.g., \texttt{subPropertyOf}) and property characteristics were not handled. 
Therefore, we only present the comparative results of the tests that could be run in both settings: ABox tests and TBox SPARQL-OWL tests.

\begin{figure}[h]
	\centering
		\includegraphics[width=1.0\textwidth]{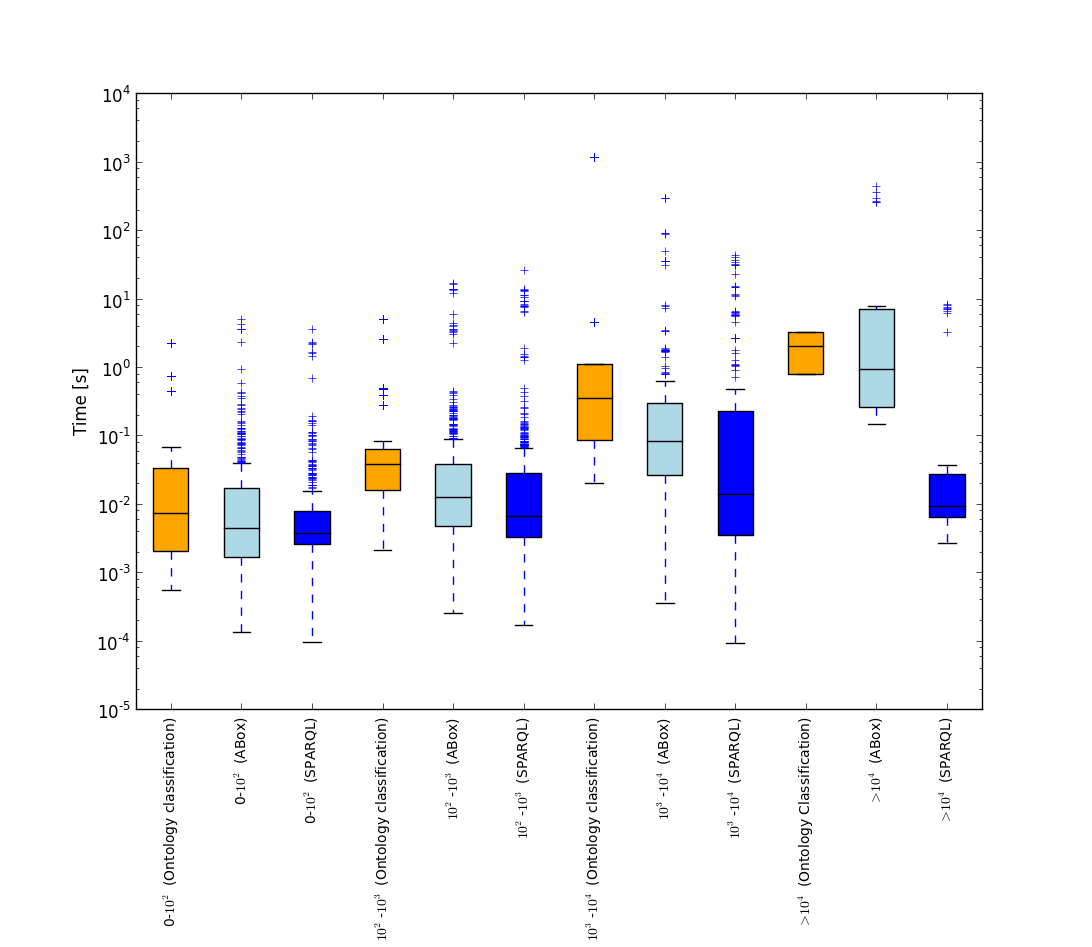}
		\caption{Test computation times per test type (ABox versus TBox based SPARQL-OWL) and ontology axiom count.}
	\label{fig:timesPerAxiomCount}
\end{figure}

The statistics are presented in the Fig \ref{fig:timesPerAxiomCount},  where X axis presents the groups of the ontologies (the ranges of the minimum and maximum number of the axioms each ontology in the group has). 
Note that the Y axis is scaled logarithmically.
On the figure, there is a box plot presenting for every group of ontologies: the median $m$ (horizontal line within the box); the first and third quartile (bottom and top line of the box); the lowest value above $m-1.5\cdot{}IQR$ (short horizontal line below the box), the highest value below $m+1.5\cdot{}IQR$ (short horizontal line above the box), where $IQR$ (interquartile range) is represented with the height of the box; and outliers (points above and below of the short lines). 

From the results as displayed in Fig. \ref{fig:timesPerAxiomCount},  it follows that TBox (SPARQL-OWL) tests are generally faster than the ABox ones, and these differences are larger in the sets of larger ontologies.
It is also apparent that the ontology classification times are large---in fact, higher on average---in comparison to the times of running the test.

\begin{figure}[h]
	\centering
		\includegraphics[width=1.0\textwidth]{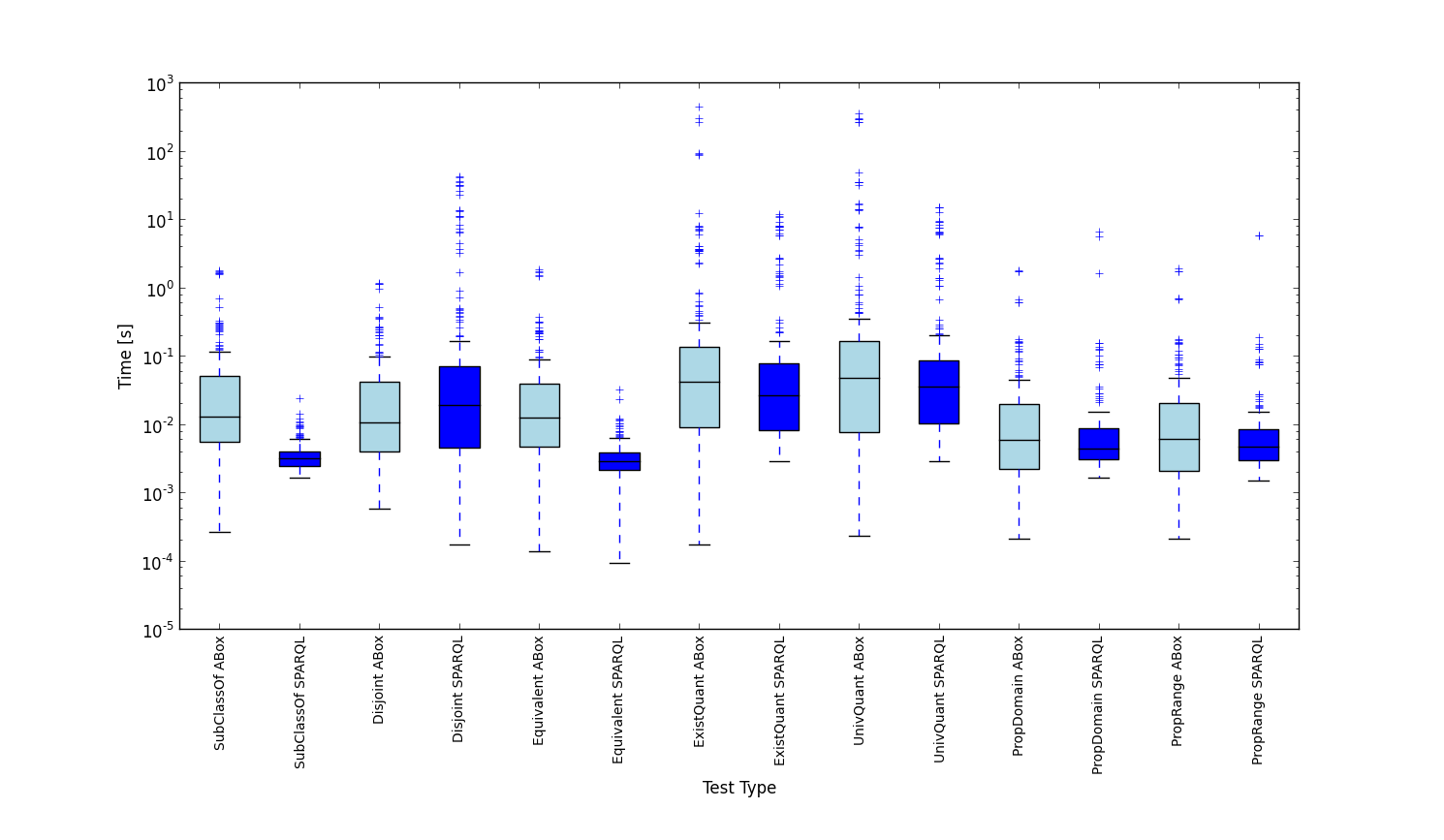}
		\caption{Test computation times per test type (ABox versus TBox based SPARQL-OWL) and per the kind of the tested axiom.}
	\label{fig:timesPerTest}
\end{figure}

In the Fig. \ref{fig:timesPerTest}, we present the running times per test type and the kind of the tested axiom.
Again, the TBox based method is generally faster, with an exception of testing disjointness.

\begin{figure}[h]
	\centering
		\includegraphics[width=1.0\textwidth]{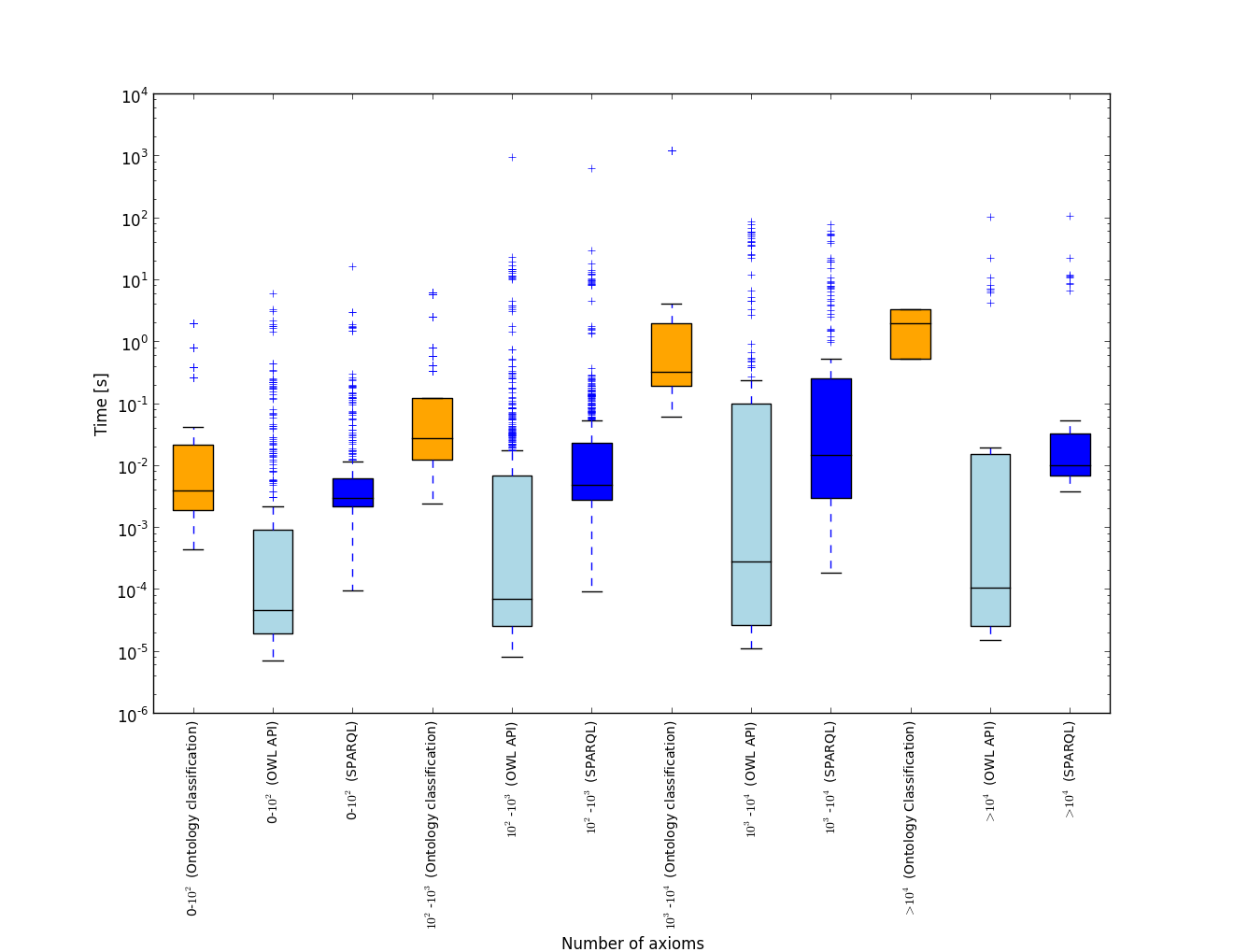}
		\caption{Test computation times per test type (OWL API versus SPARQL-OWL TBox test) and ontology axiom count.}
	\label{fig:timesPerAxiomCountTBox}
\end{figure}

\begin{figure}[h]
	\centering
		\includegraphics[width=1.0\textwidth]{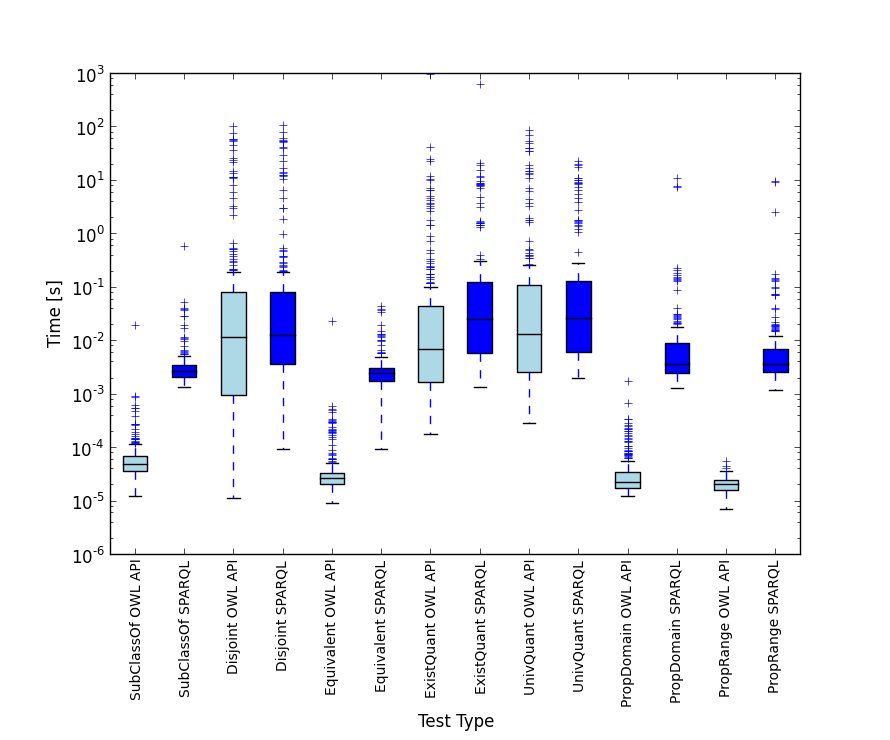}
		\caption{Test computation times per test type (OWL API versus SPARQL-OWL TBox test) and per the kind of the tested axiom.}
	\label{fig:timesPerTestTBox}
\end{figure}

We have also tested two alternative TBox querying approaches (based on SPARQL-OWL and based on using OWL API and the reasoner). 
The results of this comparison between SPARQL-OWL and OWL API are presented in Figures \ref{fig:timesPerAxiomCountTBox} and \ref{fig:timesPerTestTBox} and showing even better performance of the TBox TDD tests.

\section{Discussion}
\label{sec:disc}

One might have wondered: why not simply browse the ontology to see if the axiom is already there? There are several reasons that the browsing-way-out is not ideal. First, then one does not know the implications (unless first classified). Second, browsing large ontologies is cumbersome, and more easily results in cognitive overload that hampers speedy ontology development. Third, the browsing is purely manual checking, making it easy to overlook something. The process of finding and adding an axiom is amenable to automation that solves these three issues. Instead of manual checks and sample axioms that have to be all manually added and possibly removed during the regular `add something and lets see what happens', this can be managed in one fell swoop. In addition, the TDD tests give also the convenience and benefit of systematic regression testing. There are some hurdles to it realisation, however, which we discuss in the next subsection. Afterward, we outline how a TDD methodology for ontology engineering may look like in general terms.

\subsection{Reflections on specifying and implementing a TDD tool}

Aside from the consideration whether one would have to go the way of epistemic queries, the first main aspect was how to specify the TBox tests. For instance, some tests can be done easily with the DL query tab in Prot\'eg\'e; e.g. $T_{cs}$'s SPARQL-OWL query amounts to either:\\
\indent  {\tt C}\hfill {\em select} {\sf Super classes} \\
or, in the other direction: \\
\indent  {\tt D}\hfill {\em select} {\sf Sub classes} \\
without even having to face the consideration of having blank nodes/unnamed classes (complex class expressions) on the right-hand-side of the inclusion axiom that was not supported in Kollia et al's BGP \cite{Kollia11}. The DL Query tab interface does not have some of the required functionality, notably regarding tests for object properties, it would have to be redesigned to function as TDD test interface anyway, and it is not ideal for TDD test performance testing due to unclear times taken up by the Prot\'eg\'e interface processing. Therefore, these DL Query tab specifications have been omitted from this paper.

The more principled aspect underlying the TDD test realisation, however, is the technique to obtain the answer of a TDD test: SPARQL SELECT-queries, SPARQL-OWL's BGP that uses SPARQL answering engine and HermiT v1.3.8, SPARQL-DL with ASK queries also using the OWL API and one of the OWL 2 DL reasoners. While the difference in performance between the ABox test and TBox tests are generally explainable---the former always modifies the ontology, so requires an extra classification step---it is not for the outlier (disjointness) or why in some cases the difference is larger (subsumption, equivalence) than in others (queries with quantifiers). Further, it may be the case that overall performance may be different when a different reasoner is used, as reasoners do differ \cite{Parsia15}. Likewise, we observed a trend towards bigger differences ABox vs SPARQL testing with larger ontologies. We do not aim to address this issue here, but note it for further investigation into the matter.

A related issue is the maturity of the tools used for the performance evaluation. Of the  ontologies selected for testing, several returned errors, which were due to incompatible data types of an OWL DL ontology with OWL 2 DL-tailored tools. Further, especially querying in the context of the TDD tests for object properties faced limitations, as most of the required features were not implemented\footnote{it is, however, possible to carry out the sequence of each of the ABox test `manually' by adding the individuals, relations, run the reasoner and check the instance classification results.}. This forced us to redesign the experiment into one of `test what can be done' now and infer tendencies from that so as to have a solid, experimentally motivated, basis for deciding which technique likely will  have the best chance of success, hence, the best candidate for extending the corresponding tool. 
This appeared to be indeed preferring TBox tests over ABox tests, although most of the RBox test will have to be carried out as ABox tests.

Finally, once all test are implemented and a multi-modal interface developed to cater for the three principal use case scenarios and any other requirements emanating from the TDD methodology (see next section), user evaluations are to be conducted to evaluate whether also for ontology engineering the TDD benefits can be reaped, as observed for conceptual modelling and software development.

\subsection{A step toward a TDD ontology engineering methodology}

\begin{figure}[h]
	\centering
		\includegraphics[width=0.9\textwidth]{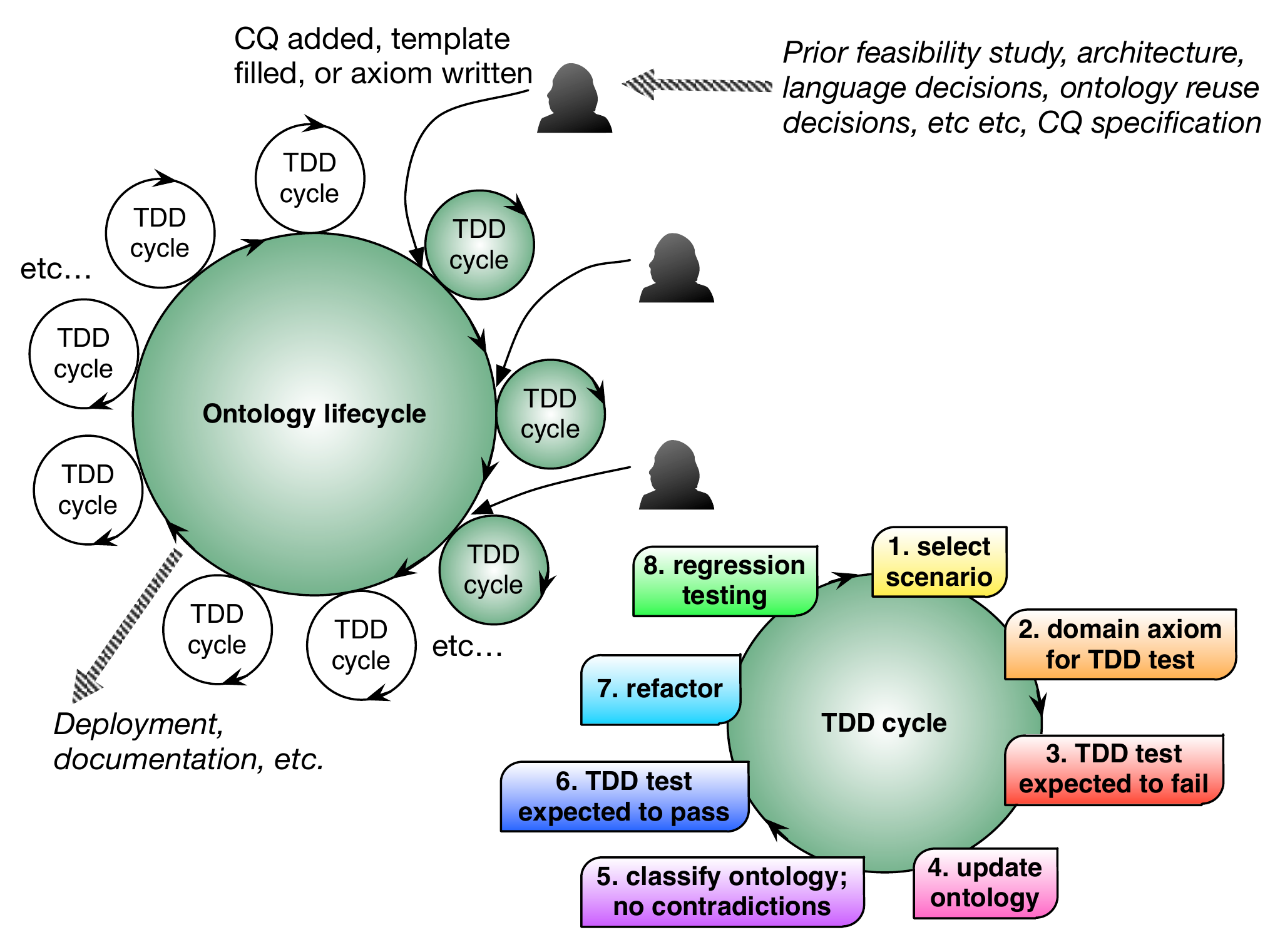}
		\caption{Sketch of a possible ontology lifecycle that focuses on TDD, and the steps of the TDD procedure summarised in key terms.}
	\label{fig:tdd-outline}
\end{figure}

A methodology is a structured collection of, principally, methods and techniques, processes, people having roles possibly in teams, and quality measures and standards across the process (see, e.g., \cite{Cockburn00}), and has been shown to improve the overall results compared to doing something without a methodology. 
This is not to say that when a full-fledged TDD ontology engineering methodology has been developed, it should be {\em the} ontology engineering methodology. Like for software engineering's `methodology selection grid' \cite{Cockburn00}, it will exceedingly suit some ontology development projects but perhaps not others. Here, with the previously described test specifications and their experimental evaluation only, we do not purport to have a full TDD methodology, but a foundational step in that direction that indicates where and how it differs in design compared to the typical waterfall, iterative, or lifecycle-based methodologies. We rework the software development TDD procedure (recall Section~\ref{sec:relwork}) into a sequence of steps applicable to ontology engineering, as follows: 
\begin{compactenum}
\item Choose the usage scenario as outlined in Section~\ref{sec:intro}: CQ-driven TDD (formalised CQ specification translated into an axiom); Authoring-driven knowledge engineer (the axiom one wants to add), or Authoring-driven domain expert (a selected template populated);
\item (Have) Select(ed) a test for the axiom;
\item Run test to check that it fails;
\item Write relevant knowledge in the ontology that should pass the test, i.e., add classes, object properties, or axioms, as applicable (this may simply be clicking a button to add the formalised CQ, the provided axiom, or the filled-in template);
\item Classify the ontology to check that nothing contradictory was added;\label{item:reason}
\item Run the same test as before, to verify it passes;
\item Optionally refactor the formally represented knowledge (including classification and checking there are no undesirable deductions, and possibly asserting the implicit knowledge);
\item Run all tests to verify that the changes to the ontology did not change the intended deductions from the ontology (regression testing); resolve any conflicting axioms or CQs (possibly by another TDD cycle).\label{item:undesdeduc}
\end{compactenum}
A sketch of the process is depicted in Fig.~\ref{fig:tdd-outline}. 
It is possible to refine these steps further, such as a way to manage the deductions following from having added new knowledge and how to handle an inconsistency or undesirable deduction due to contradictory CQs (alike resolving conflicting requirements in software engineering) that may surface in steps~\ref{item:reason} and \ref{item:undesdeduc}. These detailed aspects are left for future work.

\subsection{Answering the research questions}

The answers to the questions posed in Section~\ref{sec:intro} can be summarised as follows.

The first question was formulated rather broadly: {\em Given the TDD procedure in software engineering, then what does that mean for ontology testing when transferred to ontology development?} The main answer to this is the specification of tests for each type of axiom one can add to the ontology, which can be realised in different ways, namely, queries over the TBox and through individuals in the ABox. While the general idea is thus the same---requirement, test specification, test fails, change something, run test again to check it passes---the means of conducting a test for ontologies is thus different. One does not check code `functionality' but whether the some piece of knowledge is present and represented in the way as intended.

Regarding the second question {\em TDD requires so-called {\em mock objects} for `incomplete' parts of the code, and mainly for methods; is there a parallel to it in ontology development, or can that aspect of TDD be ignored?} can be answered in the affirmative. In particular for the general ABox tests and the so-called RBox, this `mock' thing had to be refined into mock individuals, mock classes, and mock axioms. The experimental evaluation showed this approach is not as fast as TBox tests, but there is no avoiding some ABox test especially when testing most of the object property characteristics.

Last, {\em In what way, and where, (if at all) can this be integrated as a methodological step in existing ontology engineering methodologies that are typically based on waterfall, iterative, or lifecycle principles rather than agile methodologies?} TDD ontology engineering, like TDD for software development, has its own procedure. While some aspects overlap, such as CQs (requirements), and the `formalisation' step in {\sc methontology} \cite{Fernandez99} (writing the axioms, by the knowledge engineer), both can be avoided as well: the former by the engineering, the latter by the domain expert. With some stretching of the notion of `lifecycle', the lifecycle is one of a TDD cycle only, which can be part of a larger lifecycle of multiple TDD cycles. There is no single neat `plug-in' point for TDD into the waterfall and iterative methodologies, however.

\section{Conclusions}
\label{sec:concl}

This paper introduced 36 tests for {\em Test-Driven Development} of ontologies, specifying what has to be tested, and how. Tests were specified both at the TBox-level with queries and for ABox individuals, using mock objects. The implementation of the main tests demonstrated that the TBox test approach performs better.
A high-level 8-step process for TDD ontology engineering was proposed.
Future work pertains to extending tools to also implement the remaining tests, elaborate on the methodology, and conduct use-case evaluations.

\subsubsection{Acknowledgments} This research has been supported by the National Science Centre, Poland, within the grant number 2014/13/D/ST6/02076. 

\bibliographystyle{splncs03}
\bibliography{tdd}

\end{document}